%% file: main.tex
\documentclass[a4paper, 10pt, conference]{ieeeconf}      %
\usepackage{FG2025}

\FGfinalcopy %

\IEEEoverridecommandlockouts                              %
\usepackage{graphicx} %
\usepackage{amsmath} %
\usepackage{amssymb}  %

\usepackage{accvabbrv}
\usepackage{algorithm}
\usepackage{algpseudocode}
\usepackage{xcolor,colortbl}
\usepackage{subcaption}
\usepackage{multirow}
\usepackage{booktabs}

\makeatletter
\let\NAT@parse\undefined
\makeatother
\usepackage{hyperref}
\usepackage[capitalize]{cleveref}
\crefname{equation}{}{}
\crefname{figure}{Figure}{Figures}
\crefname{table}{Table}{Tables}

\def\FGPaperID{39} %
\title{Flexible Geometric Guidance for Probabilistic Human Pose Estimation\\with Diffusion Models}

\author{\parbox{16cm}{\centering
    {\large Francis Snelgar$^{1,2}$, Ming Xu$^1$, Stephen Gould$^1$, Liang Zheng$^1$, Akshay Asthana$^2$}\\
    {\normalsize
    $^1$ School of Computing, Australian National University, Canberra, Australia\\
    $^2$ Seeing Machines, Canberra, Australia}}%
    \thanks{This work was supported by Australian Research Council (ARC) Linkage grant LP21020093.}%
}

\usepackage{fancyhdr}
\begin{document}

\ifFGfinal
\thispagestyle{empty}
\pagestyle{empty}
\else
\author{Anonymous FG2025 submission\\ Paper ID \FGPaperID \\}
\pagestyle{plain}
\fi
\maketitle
\thispagestyle{fancy}
\input{abstract}
\input{introduction}
\input{related_works}
\input{background}
\input{method}

\input{experiments}
\input{discussion}

\clearpage
\input{ethical_impact_statement}
\clearpage

{\small
\bibliographystyle{ieee}
\bibliography{references}
}

\clearpage
% \appendix
\input{supplementary}

\end{document}

%% file: abstract.tex
\begin{abstract}

3D human pose estimation from 2D images is a challenging problem due to depth ambiguity and occlusion. Because of these challenges the task is underdetermined, where there exists multiple---possibly infinite---poses that are plausible given the image. Despite this, many prior works assume the existence of a deterministic mapping and estimate a single pose given an image. Furthermore, methods based on machine learning require a large amount of paired 2D-3D data to train and suffer from generalization issues to unseen scenarios. To address both of these issues, we propose a framework for pose estimation using diffusion models, which enables sampling from a probability distribution over plausible poses which are consistent with a 2D image. Our approach falls under the guidance framework for conditional generation, and guides samples from an unconditional diffusion model, trained only on 3D data, using the gradients of the heatmaps from a 2D keypoint detector. We evaluate our method on the Human 3.6M dataset under best-of-$m$ multiple hypothesis evaluation, showing state-of-the-art performance among methods which do not require paired 2D-3D data for training. We additionally evaluate the generalization ability using the MPI-INF-3DHP and 3DPW datasets and demonstrate competitive performance. Finally, we demonstrate the flexibility of our framework by using it for novel tasks including pose generation and pose completion, without the need to train bespoke conditional models. We make code available at \url{https://github.com/fsnelgar/diffusion_pose}.

\end{abstract}

%% file: introduction.tex
\section{Introduction}

Estimating the 3D pose of a human from one or more images is a widely studied problem in computer vision with a broad range of applications including human-computer interaction, robotics and augmented reality. We highlight two key challenges in this line of work. First, estimating 3D pose using 2D images is an inherently ambiguous prediction task, caused by issues such as occlusion and monocular depth resolution. In short, the mapping from 2D images to 3D pose is underdetermined. Second, pose estimation methods based on machine learning require extensive amounts of paired 2D-3D data to learn a suitable mapping. Collecting such a dataset can require considerable annotation effort.

A promising approach to address the first challenge---one that we adopt in this paper---is to estimate a probability distribution over plausible poses, \ie, those consistent with the image data, instead of a single best pose. Given such a distribution we can then sample plausible poses and evaluate them against some downstream task. Indeed, under this framework an explicit representation of the distribution is not necessary so long as we can draw samples. %

\begin{figure}
    \centering
    \includegraphics[width=0.5\textwidth]{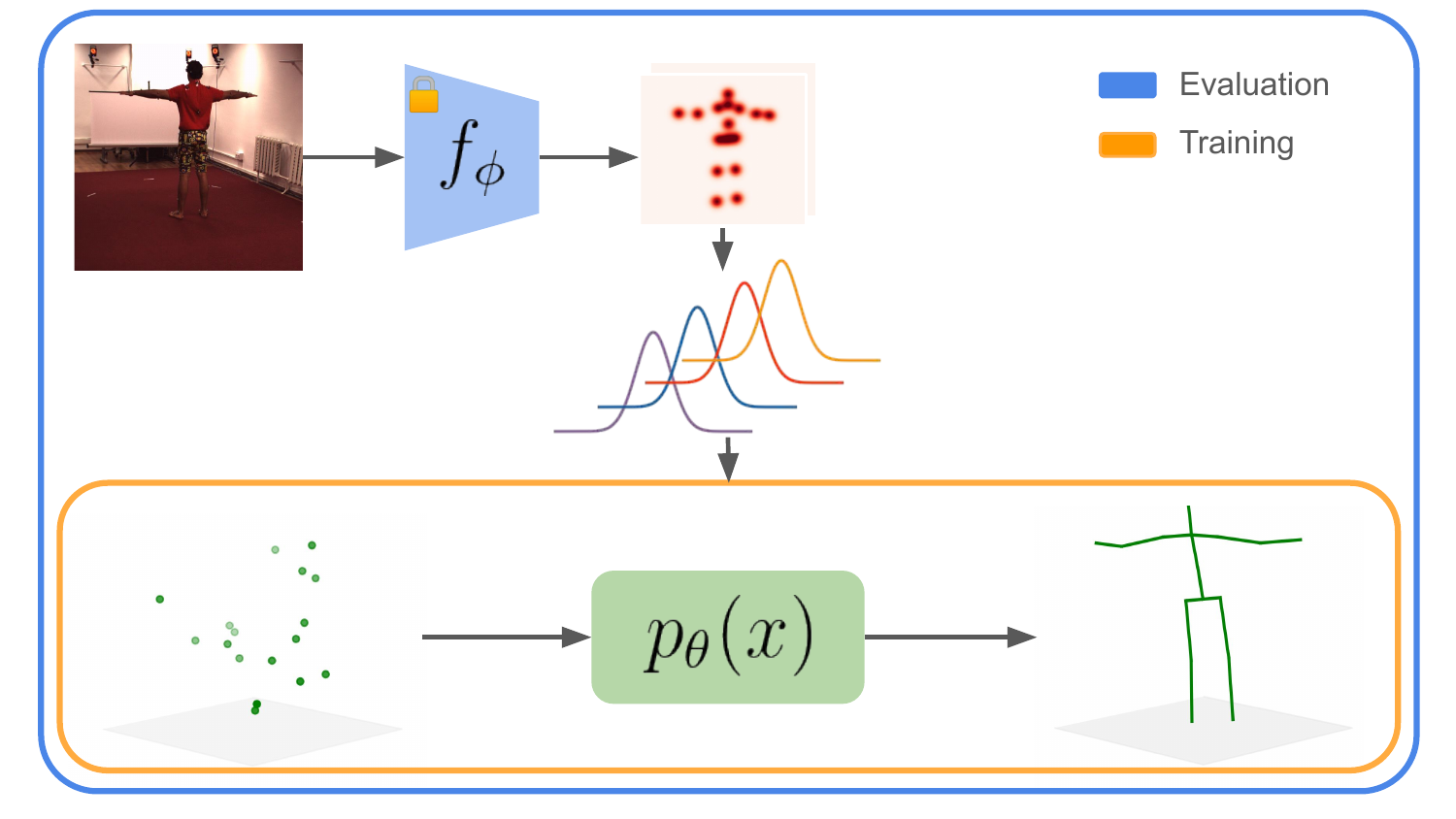}
    \caption{Overview of our pose estimation method. Using the 2D detections (modeled as Gaussians) from a keypoint detector $f_\phi$, we guide the reverse process of an unconditional diffusion model $p_\theta$ to sample 3D poses from a conditional distribution using geometric guidance. Only 3D poses are required for training.}
    \label{fig:overview}
\end{figure}

Several recent works~\cite{ci2022gfpose,li2019generating,Li2020,holmquist2022diffpose,Wehrbein2021,Jiang2024ZeDO} pursue this direction, leveraging conditional generative models. However, a limitation of these works is that they require paired datasets of conditioning signal (\ie, images) and ground truth 3D poses for training. In contrast, we \textit{guide} an unconditional human pose diffusion model using a conditioning signal which can be trained separately. The benefit of this approach is twofold. First, we can be flexible with our choice of 2D conditioning signal, \eg, different keypoint detectors or partial observation of keypoints, without retraining a new conditional generative model. Second, we introduce the idea of \textit{controllable diversity}, where we can modify the uncertainty ellipses in our 2D conditioning to vary both the magnitude and direction of diversity for conditional pose generation.

The basis of our method relies on the \textit{guidance framework}, first introduced by Dhariwal and Nicol~\cite{dhariwal2021diffusion} for class-conditional image generation. Classifier guidance has an intuitive Bayesian interpretation, where an unconditional pose diffusion model acts as a \textit{prior} distribution over plausible human poses, while the 2D conditioning signal allows for sampling from the posterior distribution of poses conditioned on the 2D input. In our work, 2D conditioning input is derived from the output of joint keypoint detectors applied to monocular RGB images.

Our contributions are as follows. First, we propose a conditional pose generation framework based on diffusion models and conditional guidance. A core capability of our approach, which improves on prior works, is decoupling the training of the 3D pose generation and 2D conditioning blocks. Second, we show how to use our framework for novel conditioning inputs such as masked joints for pose completion. Finally, we describe how to explicitly control the diversity of generated poses, all without retraining bespoke conditional diffusion models. 

%% file: related_works.tex
\section{Related Work}
In this section we present a brief overview of recent works using diffusion models. We then summarize works in the field of human pose estimation as well as recent works using probabilistic methods to address the problem.

\textbf{Diffusion models} are a recent family of generative models originally proposed by Sohl-Dickstein \etal~\cite{sohl2015deep}. They were popularized by Ho \etal~\cite{Ho20} who showed state-of-the-art results for image generation and Rombach \etal~\cite{rombach2021highresolution} and Saharia \etal~\cite{saharia2022photorealistic} showed they could successfully generate images from text conditioning. More recent works successfully applied diffusion models to domains such as human motion \cite{motionbert2022,tevet2023human}, point clouds \cite{zeng2022lion}, and 3D novel view synthesis \cite{watson2022novel}.

\textbf{2D-3D pose estimation} from images is an extensively studied task. A common approach is to assume 2D keypoint detections are available for all joints, reducing the 3D pose estimation task to a ``lifting'' problem \cite{martinez_2017_3dbaseline,pavllo:videopose3d:2019,Wehrbein2021,graformer_zhao_2022,gong2023diffpose,ci2022gfpose}. Martinez \etal~\cite{martinez_2017_3dbaseline} propose a competitive baseline method using a simple MLP network which was improved on by Zhao \etal~\cite{graformer_zhao_2022} with the use of a graph based network. Pavllo \etal~\cite{pavllo:videopose3d:2019} incorporate temporal information into the problem, further improving performance. Gong \etal~\cite{gong2023diffpose} uses a diffusion model conditioned on 2D keypoint detections. They produce a deterministic pose estimate by taking the mean of sampled 3D poses. Unlike this family of pose estimation works, our method focuses on generating a distribution of plausible 3D poses instead of a single deterministic sample.

\textbf{Correspondence free pose estimation.} 
While learning based methods have made significant progress in 2D-to-3D lifting, they require large amounts of paired 2D-3D data to train, and can suffer from cross-domain generalization issues \cite{gong2021poseaug}. Correspondence free methods require only 3D data during training, however to date their performance has lagged behind 2D-3D methods. Bogo \etal~\cite{Bogo:ECCV:2016} use the SMPL shape model \cite{SMPL:2015} to fit a mesh to detected 2D keypoints. Mueller \etal~\cite{Mueller:CVPR:2021} extended the work to better handle interpenetration and self contact in SMPLify-XMC. Gu \cite{Gu_2020} proposed a hierarchical optimization method in dynamic scenes for tracking multiple subjects. Other methods \cite{learning_to_fit, Song_2020_ECCV} use a neural network to learn a parameter update rule in a gradient descent framework. Only 3D data is required to train the update rule, and for inference the update rule is used to minimize the keypoint reprojection error.

\textbf{Probabilistic pose estimation.} Because of the ambiguity in human pose estimation introduced in part by occlusion and inaccuracies in the 2D detectors, probabilistic pose estimation has long been studied. Early works, \eg~\cite{simo-serra_2012_pose_estimation}, used stochastic sampling coupled with shape models to propagate uncertainty from the image space to the shape space and use kinematic constraints to guarantee plausible poses. Sharma \etal~\cite{Sharma_2019_ICCV} use a variational autoencoder conditioned on 2D detections and train a second network used to rank the hypotheses. Wehrbein \etal~\cite{Wehrbein2021} use a normalizing flow network conditioned on multivariate Gaussian parameters fitted to detector heatmaps to map the detector distribution to the 3D pose distribution. Diffusion models have also been applied in recent works \cite{holmquist2022diffpose,ci2022gfpose,choi2022diffupose} with the denoising model directly conditioned on detector results. These recent probabilistic works train a conditional generative model, requiring paired 2D-3D data, and are likely to over-fit to the specifics of the 2D detector used, leading to poor generalisation to new detectors. Our method has the advantage of not requiring any 2D detector or image data for training and can be explicitly guided by the characteristics of the 2D condition at evaluation time. 

Jiang \etal~\cite{Jiang2024ZeDO} also train an unconditional diffusion model to learn the prior distribution of human poses.  A key difference is that ours has a strong probabilistic interpretation through adherence to the underlying principles of DDPMs and classifier guidance by explicitly sampling from an approximation to the conditional density, whereas ZeDO use an optimization framework with a diffusion model to fix implausibility in seed poses. In particular, ZeDO is initialized with cluster centers from a nearest neighbours preprocessing step, whereas ours is initialized from a zero-mean identity-covariance Gaussian following standard DDPM theory. Furthermore, we do not constrain the joints to the rays defined by the camera center and 2D keypoints, instead allowing the observation likelihood to be weighted against the prior, and indeed in the case of pose completion, to be missing altogether. Contemporary work by Ji \etal~\cite{ji_3d_2024} is also similar to our method, however uses a truncated diffusion schedule initialized with the 2D keypoints at the ground truth depth of the pelvis joint where as our method is initialized from  $\mathcal{N}(0, \mathbf{I})$ and estimates from RootNet \cite{Moon_2019_ICCV_3DMPPE}.

\textbf{Geometric guidance in diffusion models.} Gradient based guidance for diffusion models was proposed by Dhariwal and Nichol \cite{dhariwal2021diffusion} for classifier guided image generation, where it was used in addition to a trained conditional model. Wang \etal~\cite{wang2023pd} propose a technique for camera pose estimation using a gradient guidance term to enforce epipolar geometry constraints in an iterative manner similar to Jiang \etal~\cite{Jiang2024ZeDO}. Foo \etal~\cite{Foo2023DistributionAlignedDF} apply guidance in their human mesh estimation work to enforce consistency between the predicted mesh and the 3D human pose. Similar to our gradient estimation method they also use the prediction of the denoised sample for their method. Different to our method, these geometric guidance works only apply the guidance term to part of the process and is used to complement a conditional model.

%% file: background.tex
\section{Background}

\subsection{Unconditional Generation using DDPMs}
\label{subsec:background_unconditional_generation_with_ddpm}

Denoising diffusion probabilistic models (DDPM)~\cite{Ho20,sohl2015deep} are a form of generative models that learn to generate samples from distribution $x_0 \sim p\left(x\right)$ by iteratively denoising samples taken from a simple base measure, commonly $x_T \sim \mathcal{N}\left(0, \mathbf{I}\right)$, over steps $t \in \{T, \dots , 1\}$. The so-called forward process of diffusion models is a Markov chain, where Gaussian noise is gradually added at each step as
\begin{equation}
    q\left(x_t \mid x_{t-1}\right) = \mathcal{N}\left(x_t;\sqrt{1-\beta_t}x_{t-1}, \beta_t\mathbf{I}\right),
\label{eq:forward_process_0}
\end{equation}
where $\beta_1,\dots,\beta_T$ is a schedule controlling the amount of noise added and $\alpha_t = 1-\beta_t$, $\bar\alpha_t = \prod_{i=1}^{t}\alpha_i$. As $\bar\alpha_t \rightarrow 0$, $q_t \rightarrow \mathcal{N}\left(0, \mathbf{I}\right)$.
A property of the forward process is that $x_t$ can be sampled from $x_0$ at any step $t$,
\begin{equation}
    q\left(x_t \mid x_0\right) = \mathcal{N}\left(x_t;\sqrt{\bar{\alpha}_t}x_0, (1-\bar{\alpha}_t)\mathbf{I}\right).
\label{eq:forward_process_1}
\end{equation}
The reverse process aims to recover $x_0 \sim p(x)$ from random noise $x_T \sim \mathcal{N}\left(0, \mathbf{I}\right)$. Diffusion models are trained to approximate the reverse process with a neural network $p_\theta$ with learnable weights $\theta$. %
Typically $p_\theta$ is trained to estimate the noise $\epsilon$ in $x_t$ using the simplified objective~\cite{Ho20},
\begin{equation}
    \mathcal{L}_\text{simple}(\theta) = \mathbb{E}_{t, x_0, \epsilon}\bigg[\|\epsilon - \epsilon_\theta(\sqrt{\bar{\alpha}_t}x_0 + \sqrt{1-\bar{\alpha}}\epsilon,t) \|^2\bigg].
\end{equation}

In our method, we train an unconditional pose generator using DDPMs and sample poses conditioned on 2D detections using the principled \textit{guidance} framework~\cite{dhariwal2021diffusion}, which we will describe in \cref{subsec:conditional_sampling_using_ddim}.

\textbf{Connections with score matching.} 
Dhariwal \etal~\cite{dhariwal2021diffusion, song2019generative} observe the connection of diffusion models parameterized using this noise prediction formulation to score matching methods \cite{vincent_score_matching_2011,song2020score}. Concretely,
\begin{equation}
    \nabla_{\!x_t}\log p_\theta\left(x_t\right) = -\frac{1}{\sqrt{1-\bar{\alpha}_t}}\epsilon_\theta(x_t).
\label{eq:score_function}
\end{equation}
The interpretation is interesting: each reverse process step taken by the learned model is actually taking a step in the direction of steepest ascent of the learned data density $p_\theta$. In essence, the neural network is pushing a (potentially noisy) sample $x_t$ into a region of high likelihood.

\subsection{Conditional Sampling using DDIM}
\label{subsec:conditional_sampling_using_ddim}

Dhariwal \etal~\cite{dhariwal2021diffusion} describe how to adapt score matching formulations to include a guidance term based on conditioning input. Concretely, we wish to sample from \textit{conditional distribution} $p_\theta(x \mid c)$, where $c$ is some conditioning input. Following the score matching interpretation for DDPMs, we would like to take gradient steps to maximize the conditional log-likelihood. Concretely, we would like the conditional score function $\nabla_{\!x_t}\log p_\theta\left(x_t \mid c\right)$, which after applying Bayes' rule gives 
\begin{equation}
    \nabla_{\!x_t}\log p_\theta\!\left(x_t \mid c \right) = \nabla_{\!x_t}\log p_\theta\!\left(x_t\right) + \nabla_{\!x_t}\log p\left(c \mid x_t \right),
\label{eq:classifier_guidance}
\end{equation}
noting that $p(c)$ does not depend on $x_t$. This tells us that sampling from a conditional distribution can be achieved if we can define a suitable observation likelihood for $c$. Dhariwal \etal~\cite{dhariwal2021diffusion} train a noisy classifier for class conditional image sampling. However, for pose estimation, it is straightforward to directly condition on 2D joint detections.

In practice, we find it important to scale the gradient of the observation log-likelihood by a constant factor $\gamma$. This can be interpreted as tempering (resp. sharpening) the likelihood function $p(c \mid x)$ when $\gamma \leq 1$ (resp. $\gamma > 1$).

%% file: method.tex
\section{Geometric Guidance for Pose Estimation}
\label{sec:method}

We now describe our specific formulation for DDPM and guidance for the 3D pose estimation from images task. We aim to estimate human 3D pose as a set of $J$ joints in 3D denoted by $x \in \mathbb{R}^{J \times 3}$, from a $w$-by-$h$ image, $I \in \mathbb{R}^{w\times h}$. Importantly, we do not require paired training data $(x, I)$. As standard for this task, the 3D pose $x$ is defined in the camera coordinate frame using the root relative pose definition. The root relative pose is obtained by subtracting the hip joint in the camera coordinate frame for all joints $j \in \{1,\dots,J\}$.

We train an unconditional model $p_\theta$ using the DDPM formulation introduced in \cref{subsec:background_unconditional_generation_with_ddpm} using the root relative pose representation from a dataset of only 3D poses. This gives us the ability to sample \textit{plausible} human poses.

For guidance, we assume that we additionally have access to the output of a detector giving the estimated 2D location of the $j$-th joint in pixel coordinates parameterized by a Gaussian with mean $c^{(j)}$ and covariance $\Sigma^{(j)}$. We will show later how these parameters can be estimated from a detection heatmap. Since detections are in 2D we need to project the 3D pose into the image plane using the camera intrinsic parameters~\cite{Hartley2004}. Mathematically, we have the likelihood of a 3D joint location given the image as
\begin{align}
    p(x^{(j)} \mid I) &= \mathcal{N}(\pi(x^{(j)}); c^{(j)}, \Sigma^{(j)}),
\end{align}
where $\pi$ is the 3D-to-2D projection operator and $\mathcal{N}(z; \mu, \Sigma)$ is the likelihood of a multivariate Gaussian random variable with mean $\mu$ and covariance matrix $\Sigma$.

We present an overview of the complete pipeline used for the pose estimation task in Alg.~\ref{alg:pose_estimation_pipeline}, and discuss each component in detail in the following sections.
\begin{algorithm}
    \caption{Pose Estimation Pipeline}
    \begin{algorithmic}
    \Require $p_\theta$, $T$, $c = \{c^{(j)}\}_{j=1}^{J}$, $\Sigma = \{\Sigma^{(j)}\}_{j=1}^{J}$
    \State $x_T \sim\mathcal{N}(0, \mathbf{I})$
    \For{$t$ from $T$ down to 1}
        \State $\epsilon_\theta \gets p_\theta(x_t)$
        \State $\hat{x_0} \gets\frac{1}{\sqrt{\bar{\alpha}_t}}(x_t - \sqrt{1 - \bar{\alpha}_t}\epsilon_\theta)$%
         \State $\hat{x}_0' \gets \hat{x}_0 + \gamma\nabla_{\hat{x}_0}\log p(c \mid \hat{x}_0,\Sigma)$ %
        \State $\epsilon \sim \mathcal{N}(0, \mathbf{I})$
        \State $x_{t-1} \gets \sqrt{\bar{\alpha}_t}\hat{x}_0' + (1-\bar{\alpha}_t)\epsilon$ %
    \EndFor
    \end{algorithmic}
    \label{alg:pose_estimation_pipeline}
    \end{algorithm}

\subsection{Observation Likelihood for Conditioning}

We require a model for the observation likelihood $p(c \mid x)$ for conditional sampling. This is given by our 2D detector. Specifically, for a joint $j$ where a detection is available,
\begin{equation}
\label{eq:joint_conditioning}
    p(c^{(j)} \mid x^{(j)}_t) = \mathcal{N}(c^{(j)}; \pi(x^{(j)}_t), \Sigma^{(j)}).
\end{equation}
Importantly, we do not require all joints to have an observation, and define the set of joints with valid conditioning as $\mathcal{I} \subseteq \{1, \dots, J\}$. Assuming that all $p(c^{(j)} \mid x^{(j)}_t)$ are independent, the observation likelihood is then
\begin{equation}
    p(c \mid x_t) = \prod_{j\in\mathcal{I}} p(c^{(j)} \mid x^{(j)}_t),
\end{equation}
and the gradient of the observation log-likelihood is
\begin{equation}\label{eq:obs_llh_grad}
    \nabla_{\!x_t^{(j)}}\log p(c \mid x_t) =
    \begin{cases}
     \nabla_{\!x_t^{(j)}}\log p(c^{(j)} \mid x^{(j)}_t), & j \in \mathcal{I} \\
     0, & \text{otherwise.}
    \end{cases}
\end{equation}

When results from multiple detector implementations are used, or results from different cameras there are multiple sources of observation. In this case we assume independence and simply sum the gradient contributions over the individual observation terms.

\textbf{Estimating parameters from heatmap.} When heatmap based detectors are used, the parameters of \cref{eq:joint_conditioning} can be estimated directly from the heatmap by solving a least squares problem.  Concretely, given a normalized\footnote{By normalized we mean that the elements of the heatmap sum to one.} heatmap $\mathbf{H}^{(j)} \in \mathbb{R}_{+}^{w \times h}$ we find $c^{(j)}$ and $\Sigma^{(j)}$ for each joint $j$ as
\begin{align}
    \operatorname*{argmin}_{c,\Sigma} \sum_{u=1}^{w}\sum_{v=1}^h \left\| \log H_{uv}^{(j)} - \log \mathcal{N}(y_{uv} ; c, \Sigma) \right\|_2^2,
\end{align}
where $y_{uv}$ is the 2D coordinate corresponding to the $(u,v)$-th coordinate in heatmap and $\mathcal{N}$ is the density function of a 2D multivariate Gaussian with mean $c$ and covariance $\Sigma$. 

When heatmaps are not available we use the 2D keypoint coordinates as $c^{(j)}$ and empirically choose a fixed value for $\Sigma^{(j)}$ that approximately matches values heatmap detectors are trained with \cite{Chen2018CPN}.

\textbf{Controlling diversity.} To control the diversity of the 3D poses we modify covariances of the 2D detections using $\Sigma^{(j)}$ from \cref{eq:joint_conditioning}. By modifying the eigenvalues of $\Sigma^{(j)}$ we can control the \textit{scale of diversity}, and by modifying the eigenvectors of $\Sigma^{(j)}$ we can control the \textit{axes of diversity}. We present analysis for both in \cref{subsec:experiments_controllable_diversity}.

\subsection{Gradient Estimation}
\label{subsec:gradient_estimation}

We found it beneficial in practice to set a small value for $\gamma$ ($\gamma = 2 \times 10^{-4}$ in our experiments) during the denoising process, effectively reducing the strength of the guidance relative to the pose prior. This choice is especially important in the early stages of the reverse process, since pose $x_t$ is close to random initialization and does not match the observed 2D keypoints. As a result, \cref{eq:classifier_guidance} is dominated by the conditioning term. We observed that aggressive guidance terms early in the reverse process often caused instability during denoising, leading to generation of implausible poses.

Furthermore, we found it beneficial to apply the posterior gradient update \cref{eq:classifier_guidance} to the \textit{estimated denoised sample} $\hat{x}_0$ instead of the noisy sample $x_t$ similar to \cite{bansal_universal_2023}. Recall from \cref{eq:forward_process_1} that we can estimate the denoised sample $\hat{x}_0$ from the predicted noise $\epsilon_\theta$ using the relation
\begin{equation}
    \hat{x}_0 = \frac{1}{\sqrt{\bar{\alpha}_t}}(x_t - \sqrt{1 - \bar{\alpha}_t}\epsilon_\theta).
\end{equation} 
One denoising step in our method is expressed as
\begin{equation}
    x_{t-1} \sim q(x_{t-1} \mid \hat{x}_0'),
\end{equation}
where
\begin{equation}
    \hat{x}_0' = \hat{x}_0 + \gamma\nabla_{\hat{x}_0}\log p(c \mid \hat{x}_0,\Sigma).
\end{equation}
Both heuristics are employed to improve the stability of the dynamics of the denoising process.

%% file: experiments.tex
\section{Experiments}
\label{sec:experiments}
\subsection{Datasets}

\textbf{Human 3.6M.} Human 3.6M \cite{h36m_pami} is a large scale dataset of 3.6 million images for 3D human pose estimation. It provides pose annotations from a motion capture system for four different camera views of eleven different actors performing various tasks. The dataset is split with subjects 1,5,6,7,8 for training and subjects 9 and 11 for evaluation. We follow previous works \cite{holmquist2022diffpose,Wehrbein2021} and evaluate on every $64^{th}$ frame.

\textbf{MPI-INF-3DHP.} The MPI-INF-3DHP \cite{mono-3dhp2017} evaluation dataset features six actors with a greater variation in poses than H3.6M and includes indoor, indoor green screen and outdoor scenes. Ground truth annotations are provided from a markerless motion capture system including `true' annotations and `universal' annotations compatible with the H3.6M skeleton, which has a fixed skeleton size.

\textbf{3DPW.} The 3D Poses in the Wild \cite{vonMarcard2018} evaluation dataset is a challenging dataset of diverse in-the-wild outdoor scenes captured from both static and moving cameras. It contains 60 video sequences with accurate per frame camera calibrations and 3D pose annotations obtained from video and IMU sensors.  

\subsection{Metrics}
We use the standard evaluation metrics used in 3D pose estimation literature for fair comparison against previous works.

\textbf{MPJPE.} Mean Per Joint Position Error is the mean per joint Euclidean distance between measurement and ground truth after root joint alignment.

\textbf{PA-MPJPE.} Procrustes Aligned Mean Per Joint Position Error is the mean per joint Euclidean distance between measurement and ground truth after procrustes alignment.

\textbf{PCK.} Percentage of Correct Keypoints is defined as the percentage of keypoints with Euclidean distance less than a threshold to the ground truth. We use the standard threshold of 150mm \cite{mono-3dhp2017}.

\textbf{AUC.} Area Under the Curve is the average of the PCK metric evaluated at a range of thresholds from 0mm to 150mm \cite{mono-3dhp2017}.

\subsection{Implementation}
\textbf{Diffusion model.} For the denoising network we adapt the simple baseline from Martinez \etal~\cite{martinez_2017_3dbaseline}. The network consists of blocks of linear, batch normalization, and ReLU layers repeated twice with a residual connection around each block. We use two blocks for a total of eight linear layers, and all layers have dimension 1024. The diffusion timestep is embedded using sinusoidal encoding then projected to the hidden dimension using a feed forward network. The timestep is injected into the network in every layer. The network is trained on the training set of Human 3.6M for 100,000 steps using the Adam optimizer with learning rate of $10^{-4}$. We also use exponential moving average of weights with decay rate of 0.995. The training objective is the simplified loss using a cosine $\beta$ schedule \cite{Nichol2021ImprovedDD} with 1000 steps and offset of $8 \times 10^{-3}$.

\textbf{2D conditioning information.} For Human 3.6M we use detection results from a Stacked Hourglass \cite{newell2016stacked} network pretrained on MPII \cite{andriluka14cvpr} and fine tuned on Human 3.6M provided by Ci \etal~\cite{ci2022gfpose}. For MPI-INF-3DHP and 3DPW datasets we follow previous works \cite{Jiang2024ZeDO,li2019generating} and use ground truth 2D keypoints. 

\textbf{Camera parameters.} For 2D conditioning tasks we use the camera intrinsic parameters supplied with the datasets.

\textbf{Root joint depth.} To convert the root relative pose representation into an absolute pose representation required for evaluating the observation likelihood we use the publicly available detection results from RootNet \cite{Moon_2019_ICCV_3DMPPE} for estimating the depth of the root joint.

\subsection{Multi-Hypothesis Pose Estimation}
In this section we present results for multi hypothesis 3D pose estimation using our conditional guidance framework introduced in \cref{sec:method}. We evaluate multi hypothesis performance by drawing \emph{M} samples and reporting the minimum MPJPE between all samples and the ground truth following previous works \cite{ci2022gfpose,li2019generating,Li2020,Wehrbein2021,holmquist2022diffpose, Jiang2024ZeDO}. We include methods based on conditional generation which are trained using paired 2D-3D data, as well as several recent correspondence free methods for comparison. However, note that the most relevant method for comparison is the contemporary work from Jiang \etal~\cite{Jiang2024ZeDO}, which is both probabilistic and correspondence free, similar to our method.

\textbf{Results on Human 3.6M.} 
 We evaluate on subjects 9 and 11 of the Human 3.6M dataset and present results in \cref{tab:h36m_results}. With \emph{M}=50, our method is comparable to conditional generation models which are trained using large amounts of paired 2D-3D data. Notably, we improve on the existing state-of-the-art correspondence free probabilistic method by Jiang \etal \cite{Jiang2024ZeDO}.

\begin{table}[]
    \centering
    \caption{Pose estimation performance on the Human 3.6M dataset. \textit{Paired 2D-3D} indicates that the method requires 2D-3D correspondences for training while \textit{3D Only} indicates only 3D data is required. For probabilistic methods \textit{M} indicates the number of hypotheses used. Keypoint detections from 2D keypoint detectors such as HRNet \cite{rombach2021highresolution} or Stacked Hourglass Network \cite{newell2016stacked} are used.}
    \input{tables_h36m_results.tex}
    \label{tab:h36m_results}
\end{table}

\textbf{Results on MPI-INF-3DHP.} We evaulate the cross domain generalization of our method on the MPI-INF-3DHP dataset. The pose prior model is trained on the Human 3.6M dataset and evaluated on MPI-INF-3DHP without additional fine tuning. We report results for \emph{M}=50 in \cref{tab:mpii_3dhp_results}. Our method performs well in the cross domain setting and is comparable with current state-of-the-art correspondence free methods.

\begin{table}[]
    \centering
    \caption{MPI-INF-3DHP results. Results marked with * are taken from \cite{Jiang2024ZeDO}. CD indicates if the evaluation is cross domain, \ie, methods were not fine tuned on the 3DHP dataset.}
    \input{tables_3dhp_results.tex}
    \label{tab:mpii_3dhp_results}
\end{table}

\begin{figure}
    \centering
    \includegraphics{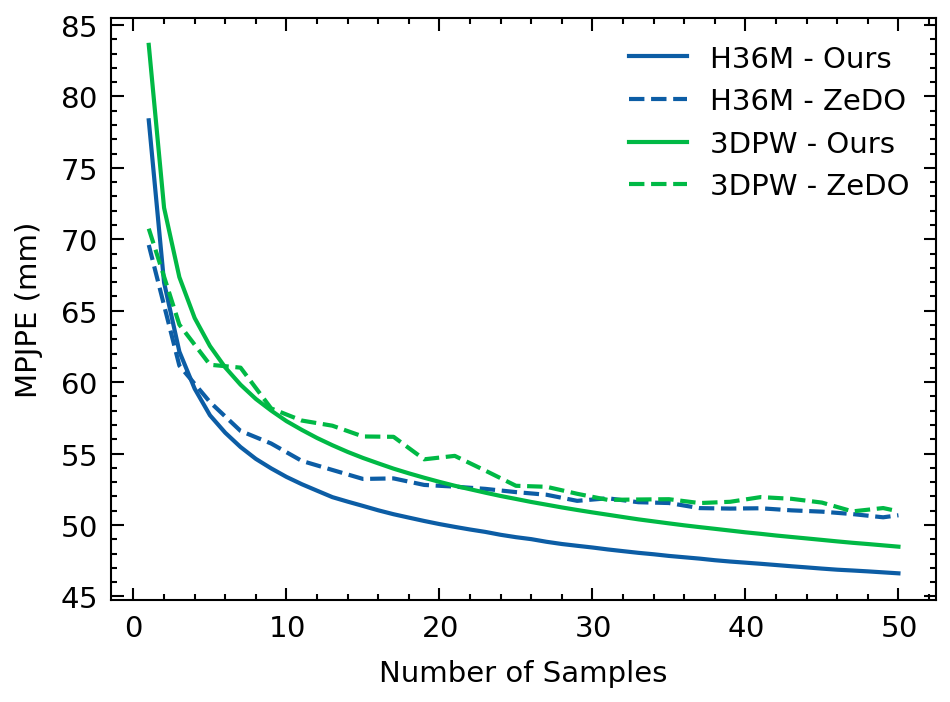}%
    \caption{MPJPE as a function of number of samples for Human 3.6M and 3DPW datasets. Results for ZeDO \cite{Jiang2024ZeDO} were reproduced from the official repository.}
    \label{fig:number_of_hypo}
\end{figure}

\textbf{Results on 3DPW.} The 3DPW dataset is a particularly challenging in-the-wild dataset with diverse poses from outdoor scenes. We  follow the same cross-domain evaluation protocol and use the pose prior model trained on Human 3.6M without fine tuning for evaluation. Results are presented in \cref{tab:3dpw_results}. While performance of our method with \emph{M}=1 lags recent work, when increased to \emph{M}=50 we improve on current state-of-the-art, highlighting the increased diversity (but plausible) poses generated by our method.

\begin{table}[]
    \centering
    \caption{Results on the 3DPW dataset. Results marked with * are taken from \cite{Jiang2024ZeDO} and results marked with $\dagger$ were produced using code from the official repository. CD indicates if the evaluation is cross domain, \ie, methods were not fine tuned on the 3DPW dataset.}
    \input{tables_3dpw_results.tex}
    \label{tab:3dpw_results}
\end{table}

\textbf{Analysis of number of hypotheses.} Because the monocular pose estimation task is underdetermined, we aim to estimate a probabilistic distribution of poses. While the performance of our method lags current state-of-the-art for a single deterministic hypothesis, we present the effect of increasing the number of hypotheses in \cref{fig:number_of_hypo} and compare the trade off between single hypothesis and multiple hypothesis performance for our method and Jiang \etal~\cite{Jiang2024ZeDO} on the Human 3.6M and 3DPW datasets. The trend across both datasets is for our methods performance to lag behind for small number of hypotheses, but continues to improve as the number increases while Jiang \etal method \cite{Jiang2024ZeDO} begins to plateau. Also note that it is trivial for our method to increase the number of hypotheses, whereas Jiang \etal~\cite{Jiang2024ZeDO} requires k-means clustering for a given value of \emph{M}, with each hypothesis initialized from a different cluster centroid.

\subsection{Flexible Generation}
We additionally demonstrate the flexibility of our method by applying it to different tasks using the same pretrained model.

\begin{figure}
    \centering
    \includegraphics{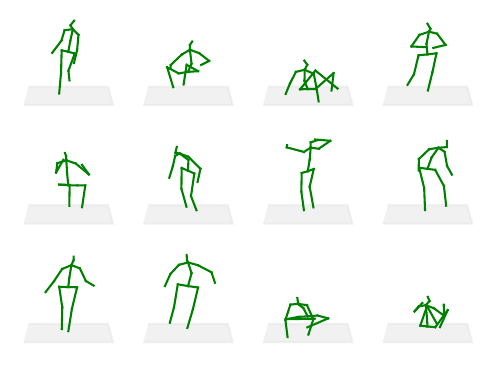}
    \caption{Qualitative examples of diverse 3D human poses generated by our method without guidance.}
    \label{fig:pose_generation_examples}
\end{figure}

\textbf{Effectiveness on unconditional 3D pose generation.} Without guidance our denoising network acts as a pose generator, drawing samples from Gaussian noise. We show several qualitative examples in \cref{fig:pose_generation_examples}. The samples are visually plausible and exhibit diversity between samples, indicating our model has successfully learned the distribution of plausible human poses.
\begin{figure}
    \centering
    \includegraphics{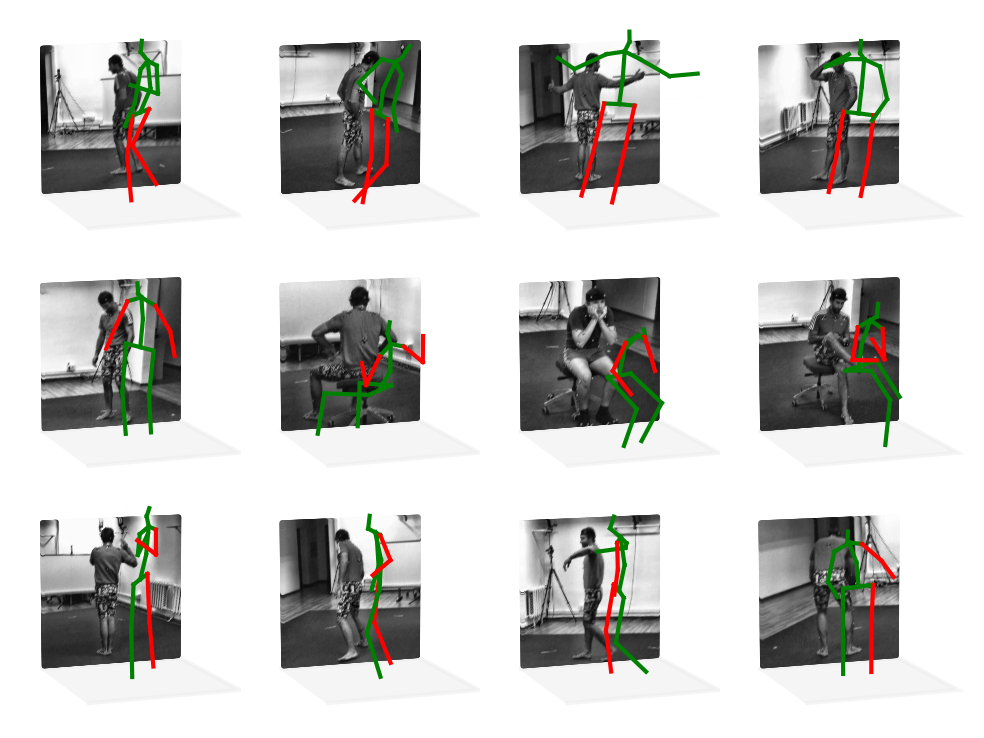}
    \caption{Qualitative examples of pose completion results. Observation likelihood is not defined for red joints and the model must `inpaint' these joints. Grayscale images illustrated for context.}
    \label{fig:joint_competion_examples}
\end{figure}

\textbf{Effectiveness on pose completion.} In pose estimation tasks it is common for keypoints to be occluded in the image and it is necessary for methods to be able to estimate poses given incomplete detection results. We evaluate the quality of our learned pose prior by applying it to a pose completion task. This task simulates occlusion by removing the condition for different subsets of joints and evaluates the ability of methods to recover plausible poses when conditioned on incomplete detection results. We remove the observation likelihood for missing joints, requiring the pose prior to inpaint plausible completions. For joints that are not missing, we use the observation likelihoods previously described. We show selected examples of pose completion results in \cref{fig:joint_competion_examples}.
Note that while the masked joints may not exactly match the image, they are consistent with the rest of the joints, and as a complete pose the results are plausible. We observe in particular that the range of movement in the arms is particularly diverse, which aligns with natural human motion.

\subsection{Controllable Diversity}
\label{subsec:experiments_controllable_diversity}
A significant advantage of our method is that through the use of different likelihoods, it is possible to tailor the level of diversity as required by the application. In the following section we evaluate the diversity of the generated poses, and the impact of different observation likelihood functions on the diversity.

\begin{figure}
    \centering
    \includegraphics{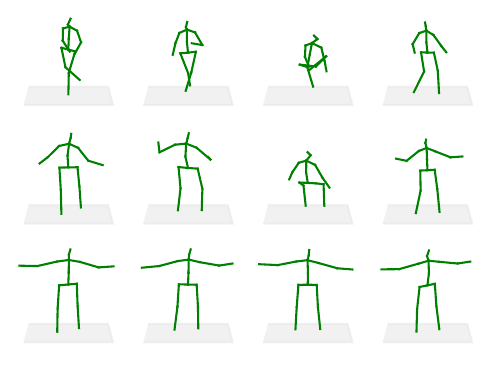}
    \caption{ Qualitative examples of the effect of scaling covariance matrices $\Sigma^{(j)}$. Each column has the same latent variable; the scaling factor decreases down the column.}
    \label{fig:joint_variation_examples}
\end{figure}

\begin{figure}
    \centering
    \includegraphics{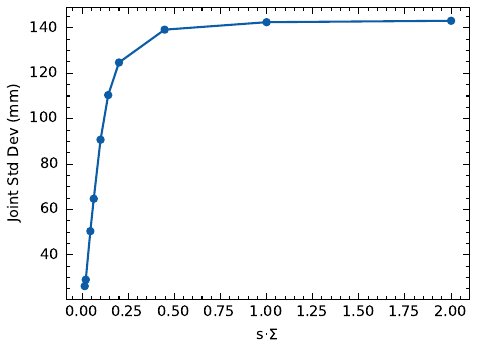}
    \caption{Effect of applying a scaling factor $s$ to $\Sigma$ on pose diversity. The mean per joint standard deviation for 200 samples is reported for different values of $s$.}
    \label{fig:pose_variation}
\end{figure}

\textbf{Diversity magnitude.} The Gaussian likelihood function is parameterized by covariance matrices $\Sigma^{(j)}$. By scaling these matrices by a constant factor $s$, it is possible to control the \textit{magnitude} of diversity. We demonstrate this capability by evaluating the impact on the multi hypothesis pose estimation task using the Human-3.6M dataset. We present qualitative examples for different values of $s\cdot\Sigma^{(j)}$ in \cref{fig:joint_variation_examples}. For small values of $s$, generated poses are more uniform and are consistent with the condition, with diversity increasing for larger values of $s$, making it possible to trade off diversity and plausibility as the application requires. To verify this observation we measure the mean per joint standard deviation of \emph{M} samples, and present the results in \cref{fig:pose_variation}. There is a clear trend showing more variation with larger values of $s$, with performance saturating as the method approaches unconditional generation.

\begin{figure*}
    \includegraphics[width=\textwidth]{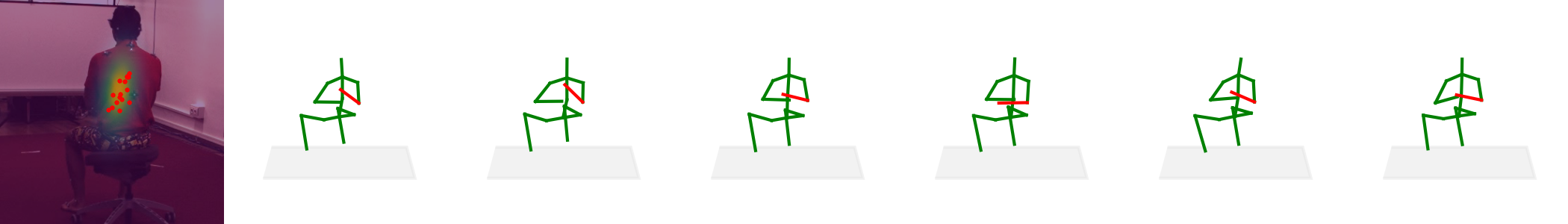}
    \includegraphics[width=\textwidth]{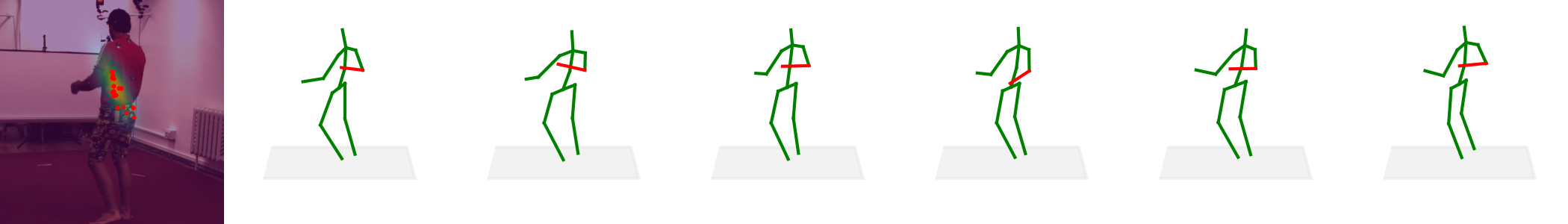}
    \includegraphics[width=\textwidth]{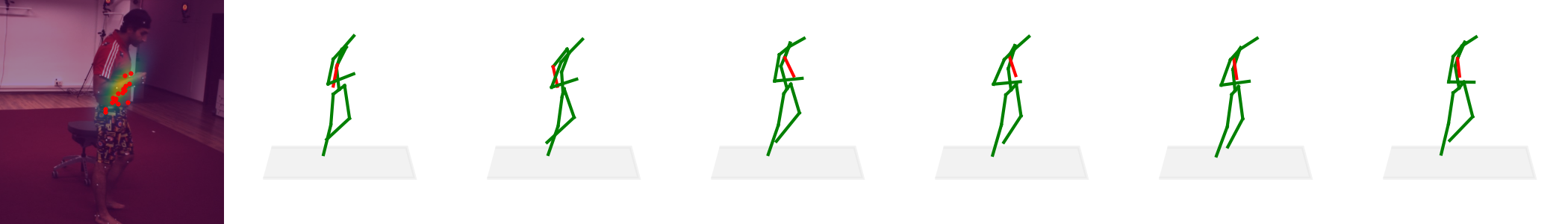}
    \caption{Qualitative examples for high 2D detector uncertainty. Overlays of the detector heatmap and 2D projection of sampled poses are shown on the left. Only the joint with high uncertainty is shown for clarity. Panels to the right contain corresponding 3D pose results. Notice the heavy occlusions present in these images, which highlights the inherent ambiguity of the task.}
    \label{fig:heatmap_examples}
\end{figure*}

\textbf{Diversity axes.} While modifying the scale of the covariance matrices $\Sigma^{(j)}$ changes the \textit{magnitude} of diversity, we also demonstrate that our method allows control of the \textit{axes} of diversity through rotating and stretching of $\Sigma^{(j)}$. We observe that this occurs naturally in heatmap based detectors when occlusion is present, with the heatmaps becoming large and rotated. This is captured by the Gaussian parameterization of the observation likelihood. We present qualitative examples of different covariances in \cref{fig:heatmap_examples} of cases where this occurs. Visually the heatmaps capture the uncertainty in the true pose in the image plane, and the projection of the occluded joint from the drawn samples is distributed along the major axis of the heatmap, mirroring the uncertainty from the detector. Note that the projections in the second row appear multi modal, which may indicate that not all poses that are consistent with the observation likelihood agree with the learned distribution of the pose prior.

%% file: tables_h36m_results.tex
\setlength{\tabcolsep}{5pt}
\begin{tabular}{l|cccc}
\toprule
Data & Method & M & MPJPE $\downarrow$ & PA-MPJPE $\downarrow$ \\
\midrule
\multirow{8}{*}{\rotatebox[origin=c]{90}{Paired 2D-3D}} & Martinez \etal~\cite{martinez_2017_3dbaseline} & & 62.9 & 47.7\\
& Gong \etal~\cite{gong2023diffpose} & 5 & 49.7 & 31.6\\
& Li \etal~\cite{Li2020} & 10 & 73.9 & 44.3\\
& Li \etal~\cite{li2019generating} & 5 & 52.7 & 42.6\\
& Sharma \etal~\cite{Sharma_2019_ICCV} & 200 & 46.8 & 37.3\\
& Wehrbein \etal~\cite{Wehrbein2021} & 200 & 44.3 & 32.4\\
& Holmquist \etal~\cite{holmquist2022diffpose} & 200 & 43.3 & 32.0 \\
& Ci \etal~\cite{ci2022gfpose} & 200 & 35.6 & 30.5 \\

\midrule
\multirow{8}{*}{\rotatebox[origin=c]{90}{3D Only}} & Gu \cite{Gu_2020} &  & 77.2 & - \\
& Fan \etal~\cite{fan2021revitalizing} & & 61.5 &  48.2\\
& Bogo \etal~\cite{Bogo:ECCV:2016} & & 82.3 &  -  \\
& Song \etal~\cite{Song_2020_ECCV} & & - & 56.4 \\
& Jiang \etal~\cite{Jiang2024ZeDO} & 1 & 65.7 & 49.0 \\
& Jiang \etal~\cite{Jiang2024ZeDO} & 50 & 51.4 & 42.1 \\
& Ours & 1 & 78.6 & 58.2 \\ %
& Ours  & 50 & 46.3 & 37.3 \\ %
& Ours  & 200 & 42.3 & 34.4\\ %

\bottomrule
\end{tabular}

%% file: tables_3dhp_results.tex
\setlength{\tabcolsep}{5pt}
\begin{tabular}{l|llllll}
\toprule
Data & Method & CD & M & MPJPE $\downarrow$ & PCK $\uparrow$ & AUC $\uparrow$\\
\midrule
\multirow{6}{*}{\rotatebox[origin=c]{90}{Paired 2D-3D}} 
& Martinez \etal~\cite{martinez_2017_3dbaseline}*   &           &     & 84.3 & 85.0 & 52.0 \\
& Ci \etal~\cite{ci2022gfpose}                      &\checkmark & 200 & -    & 86.9 & - \\
& Gholami \etal~\cite{Gholami_2022_CVPR}            &\checkmark &     & 77.2 & 88.4 & 54.2  \\
& Gong \etal~\cite{gong2021poseaug}                 &\checkmark &     & 73.0 & 88.6 & 57.3 \\
&&&&&\\
&&&&&\\
\midrule
\multirow{4}{*}{\rotatebox[origin=c]{90}{3D Only}} 
& Jiang \etal~\cite{Jiang2024ZeDO}  &\checkmark & 50 & 69.9 & 90.2   & 58.8 \\
& Ours                              &\checkmark & 50 & 73.2 & 89.1   & 56.2\\ %
&&&&&\\
&&&&&\\

\bottomrule
\end{tabular}

%% file: tables_3dpw_results.tex
\setlength{\tabcolsep}{5pt}
\begin{tabular}{l|lllll}
\toprule
Data & Method & CD & M & MPJPE $\downarrow$ & PA-MPJPE $\downarrow$ \\
\midrule
\multirow{5}{*}{\rotatebox[origin=c]{90}{Paired 2D-3D}} 
& Ma \etal~\cite{Ma_2023_CVPR}              &           &   & 67.5  & 41.3 \\
& Gong \etal~\cite{gong2021poseaug}*        &\checkmark &   & 94.1  & 58.5 \\
& Gholami \etal~\cite{Gholami_2022_CVPR}    &\checkmark &   & 81.2  & 46.5 \\
&&&&&\\
&&&&&\\

\midrule
\multirow{6}{*}{\rotatebox[origin=c]{90}{3D Only}} 
& Song \etal~\cite{Song_2020_ECCV}              &           &       & -     & 55.9 \\
& Fan \etal~\cite{fan2021revitalizing}          &           &       & 98.6  & 68.0 \\
& Jiang \etal~\cite{Jiang2024ZeDO}              &\checkmark & 1     &  69.7 & 40.3 \\
& Jiang \etal~\cite{Jiang2024ZeDO}$\dagger$     &\checkmark & 50    &  54.8 & 30.6 \\
& Ours                                          &\checkmark & 1     & 79.9  & 50.1 \\
& Ours                                          &\checkmark & 50    & 48.5  & 30.4 \\
\bottomrule
\end{tabular}

%% file: discussion.tex
\section{Further Analysis}
In this section we present an ablation study on the components of our pose estimation system. 

\textbf{Reverse Process Gradients:} We analyze the impact of the alternate update formulation introduced in \cref{subsec:gradient_estimation} using the 3DPW dataset. Results are presented in \cref{tab:ablation}. Our proposed method substantially improves performance which we attribute to two reasons. First, $\nabla\text{p}(c \mid x_t)$ is noisy with large magnitude early in the reverse process when $x_t$ is close to random noise, which can push samples out of the learned distribution causing implausible samples. Second, as $t \rightarrow 0$,  $\bar{\alpha}_t \rightarrow 1$, and the classifier guidance update $\sqrt{1-\bar{\alpha}_t}\nabla\log\text{p}(c \mid x_t) \rightarrow 0$, providing minimal guidance to the reverse process. In contrast, our gradient update $\gamma\nabla\log\text{p}(c \mid \hat{x}_0)$  has a constant scaling factor $\gamma$, and provides stable and consistent dynamics throughout the reverse process.

\textbf{Root Joint:} As the generative pose prior operates in root-relative coordinates, we use RootNet \cite{Moon_2019_ICCV_3DMPPE} to predict the 3D root joint position. We assume error in these predictions and for probabilistic estimation we sample around these predictions. Concretely, given a RootNet model $f\phi$, we sample root joint positions $x^{root} \sim \mathcal{N}(f_\phi(I), \Sigma)$ with a small fixed variance $\Sigma$. We present the impact of root joint positions in \cref{tab:ablation}. Sampling around RootNet predictions slightly improves performance, while using the ground truth 3D position further improves performance.
\begin{table}[]
    \centering
    \include{tables_3dpw_ablation.tex}
    \caption{Ablation analysis for different configurations of our method. Results are for multiple hypothesis pose estimation on the 3DPW dataset using 50 samples. Grad Update indicates the use of the gradient update introduced in \cref{subsec:gradient_estimation}, Root Joint Sampling indicates sampling around the detected root joint position, GT Root Joint indicates the use of the ground truth root joint position.}
    \label{tab:ablation}
\end{table}

\textbf{2D Keypoint Detector:} In \cref{sec:experiments} we follow Jiang \etal~\cite{Jiang2024ZeDO} and use Stacked Hourglass \cite{newell2016stacked} keypoints for Human 3.6M dataset and ground truth keypoints for MPI-INF-3DHP and 3DPW datasets. We present additional results in \cref{tab:detector_ablation} using HRNet \cite{sun2019deep} for completeness. These datasets are more diverse than Human 3.6M with outdoor and in-the-wild scenes featuring multiple subjects, and without fine tuning, keypoint detection accuracy lags behind.%

\begin{table}[]
    \centering

\include{tables_detector_ablation.tex}
    \caption{Multiple hypothesis pose estimation using ground truth and detected keypoints for all datasets with 50 samples. GT indicates if ground truth keypoints were used. Results marked with $^\star$ are taken from \cite{Jiang2024ZeDO}, those marked with $^\dagger$ were produced using the official repository. }
    \label{tab:detector_ablation}
\end{table}

\section{Conclusion}

In this work, we present a novel and flexible geometric guidance framework for probabilistic human pose estimation based on principled guidance theory. Our framework decouples 3D pose generation and 2D detection, alleviating the need for training sets of paired 2D-3D data. We show state-of-the-art correspondence free performance for probabilistic human pose estimation on the Human 3.6M dataset, and competitive performance on the MPI-INF-3DHP  and 3DPW datasets. We demonstrate the flexibility of our method by showing that our human pose prior can be used for unconditional generation and propose extending the guidance framework for pose completion tasks, all without the need to train bespoke conditional models.

\textbf{Limitations and future works.}  A limitation of our method is that DDPMs require repeated sampling during the reverse process, which may not be practical for real time applications. To address this, investigating faster sampling methods such as DDIM \cite{song2020score} would be a promising direction. The best-of-m method used for evaluating probabilistic models is a current limitation for in-the-wild evaluation as it requires ground truth annotations, which should be addressed in future works. Our geometric guidance uses a simple Gaussian model for the observation likelihood, and does not take advantage of cues such as temporal consistency. An interesting direction for future work could be to extend our method to utilize more expressive observation likelihoods and furthermore, incorporate temporal information from videos.

%% file: tables_3dpw_ablation.tex
\setlength{\tabcolsep}{5pt}
\begin{tabular}{c|c|c|c}
\toprule
Grad Update & Root Joint Sampling & GT Root Joint & MPJPE $\downarrow$ \\
\midrule
& & & 112.8 \\
\checkmark & & & 53.1 \\
\checkmark & \checkmark & & 48.5 \\
\checkmark & & \checkmark & 43.6
\end{tabular}

%% file: tables_detector_ablation.tex
\setlength{\tabcolsep}{5pt}
\begin{tabular}{l|c|ccc}
\toprule
& & \multicolumn{3}{c}{MPJPE $\downarrow$} \\
Method & GT & H36M & 3DHP & 3DPW \\
\midrule
Jiang \etal~\cite{Jiang2024ZeDO} & \checkmark &35.7$^\dagger$  & 69.9$^\star$ &  54.8$^\star$  \\
 & & 51.4$^\star$ & 115.9$^\dagger$ & 105.4$^\dagger$ \\
Ours & \checkmark & 36.8 &  73.2 & 48.5  \\
 & & 46.3 & 115.2 & 92.3 \\
\end{tabular}

%% file: ethical_impact_statement.tex
\section{Ethical Impact Statement}

\textbf{Risks:}
As our method deals with human pose estimation, the use of subjects biometric data, and their consent to this use must be considered. We use both videos of subjects and their 3D poses in the training and evaluation of our method, which introduces the possibility of private biometric data being captured or memorized in the models we train. Additionally the statistical distribution of the training data should be considered, as it introduces the risk of our method being unfairly biased against particular demographic or human behaviors.

\textbf{Strategies:}
All three datasets we use in our experiments (H3.6M \cite{h36m_pami}, 3DHP \cite{mono-3dhp2017} and 3DPW \cite{vonMarcard2018}) are datasets captured specifically for the purposes of scientific research using actors who have consented to their data being captured for this purpose. To mitigate bias in methods using the data, the datasets are constructed with using a number of different actors (H36M: 11, 3DHP: 8, 3DPW: 7) of both genders (H36M: 5/6, 3DHP: 4/4), and performing a wide range of different tasks (H36M: 15, 3DHP: 8) and `in-the-wild' sequences (3DPW). 
Additionally, our method never combines both video and biometric modalities into a single model, with the keypoint detector only trained with video data, while the pose diffusion model is only trained with 3D pose data.

\textbf{Benefit Risk Analysis:}
While there is always a risk of machine learning models memorizing biometric data, we mitigate this risk through the use of public datasets of actors who have given their consent. Additionally our method effectively eliminates the chances of multiple modalities of data being memorized in the single model due to the decoupled nature of our method. The authors of the datasets attempt to minimize potential bias in the data, however the variation of subjects and behaviors is relatively small and models trained using these datasets are likely to suffer from bias of some form. As our method is to be used for academic purposes only, and not in any safety critical capacity, the potential side effects of potential bias is limited to poor generalization and therefore is minor.

%% file: supplementary.tex
\section{2D Keypoint Alignment} 

\begin{figure}
    \centering
    \includegraphics{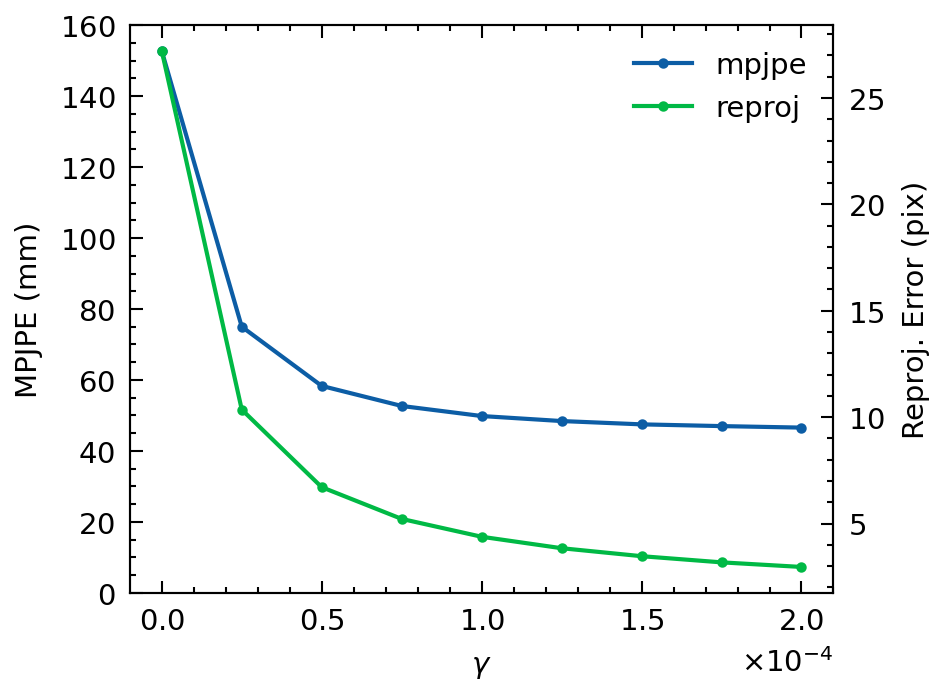}
    \caption{MPJPE (blue) and reprojection error (green) as a function of guidance scale $\gamma$.}
    \label{fig:h36m_gamma_vs_kp_reproj}
\end{figure}

\begin{figure}
    \centering
    \includegraphics{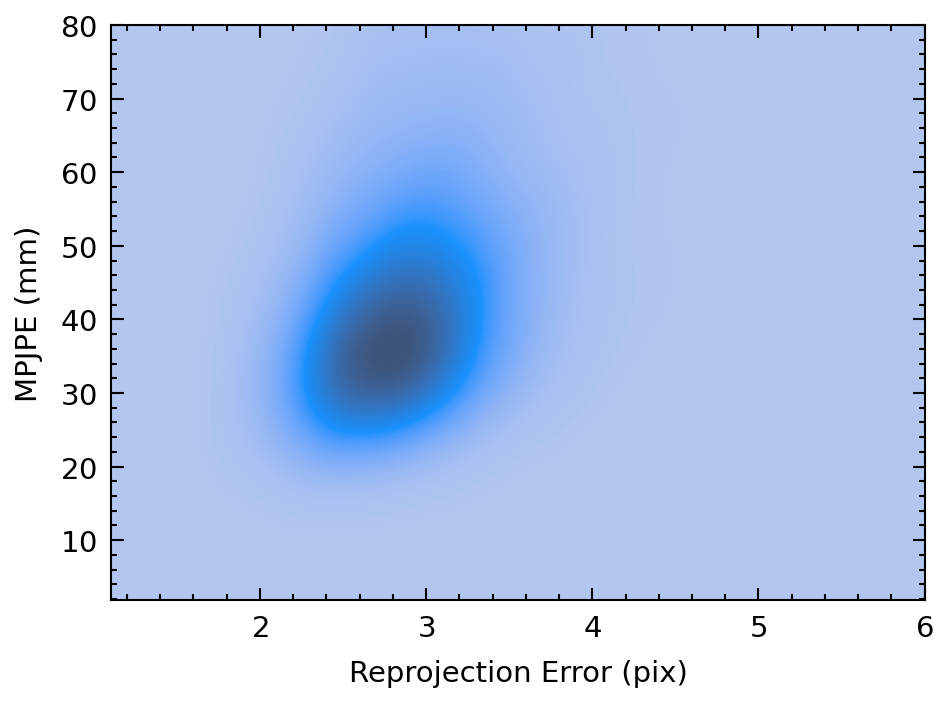}
    \caption{Joint distribution of MPJPE and reprojection error for $\gamma = 2\times10^{-4}$.}
    \label{fig:h36m_kp_reproj_error}
\end{figure}

For the multiple hypothesis pose estimation experiments in  \cref{sec:experiments} we use a fixed guidance scale $\gamma = 2\times10^{-4}$. Intuitively, the guidance scale provides a trade off between the observation likelihood and the pose prior. To quantify the alignment of generated poses with the observation likelihood we plot the reprojection error between generated poses and the 2D keypoints in \cref{fig:h36m_gamma_vs_kp_reproj}. We observe that error decreases as $\gamma$ increases, i.e, generated poses are better aligned with the observation likelihood. We additionally show the joint distribution of reprojection error and MPJPE in \cref{fig:h36m_kp_reproj_error} and observe that the distribution is weakly correlated, suggesting that for a large enough $\gamma$ ($2\times10^{-4}$) the reprojection error is minimal and misalignment with the observation likelihood does not significantly contribute to MPJPE.

\section{Keypoint Detector Failure Cases}
\begin{figure*}
    \includegraphics[width=\textwidth]{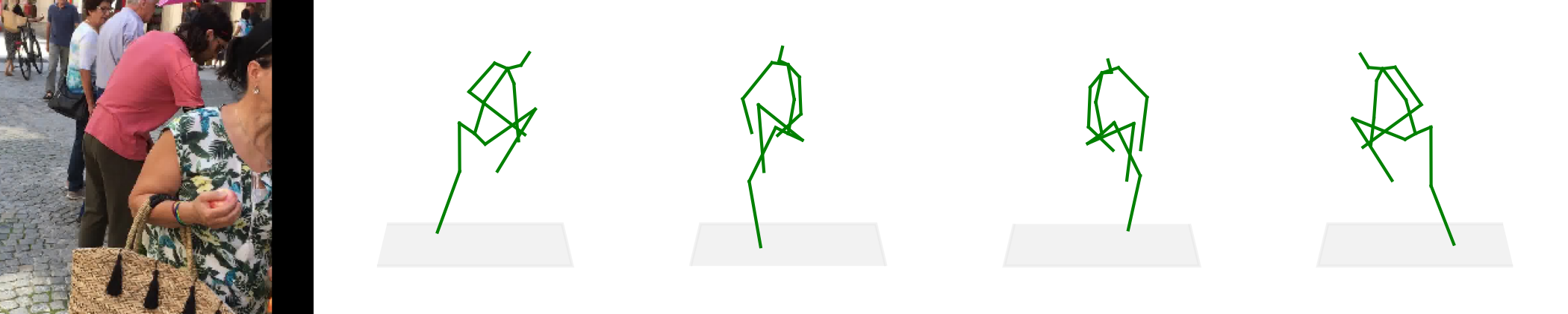}
    \includegraphics[width=\textwidth]{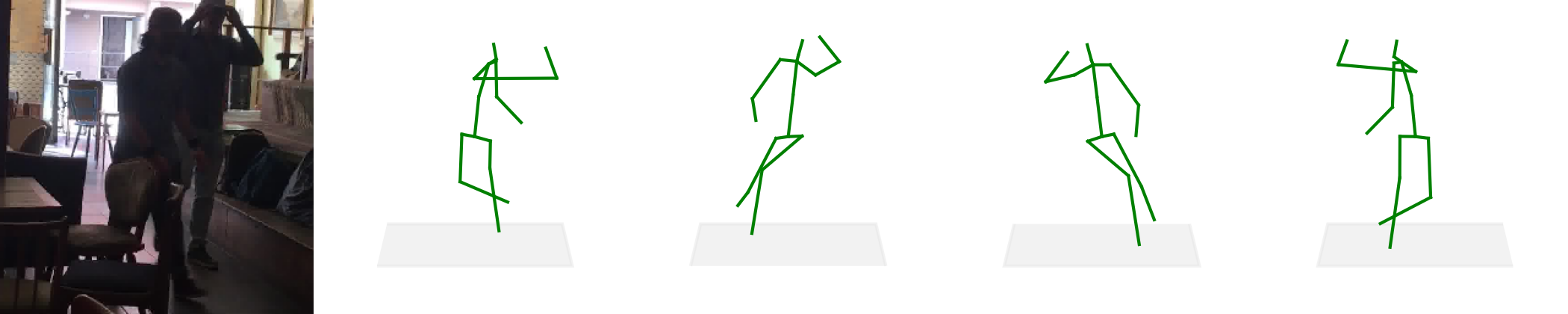}
    \includegraphics[width=\textwidth]{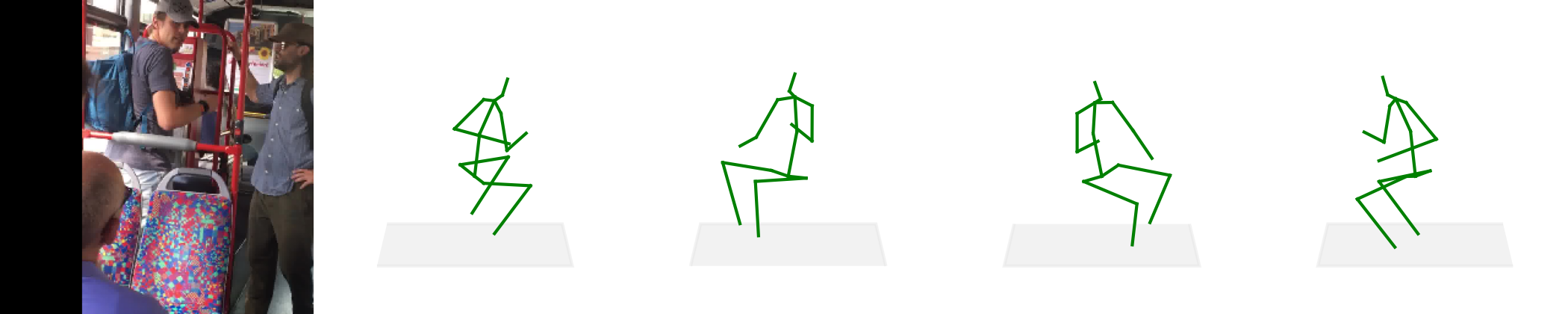}
    \caption{Failure cases for pose estimation on the 3DPW dataset. Poses are shown from the front, left, right and rear view points in each column. Rows one and three show keypoint detection failure due to occlusion, row two shows failure due to detection of wrong subject.}
    \label{fig:failure_cases_3dpw}
\end{figure*}

We present qualitative example of pose estimation inaccuracy caused by keypoint detection failure in \cref{fig:failure_cases_3dpw}. In the first and third rows, occlusion of the legs cause keypoint detection failure, and in the second row, keypoints from the background subject have been incorrectly assigned to the foreground subject.

\section{Multiple Hypothesis Examples}
\begin{figure*}
    \includegraphics[width=\textwidth]{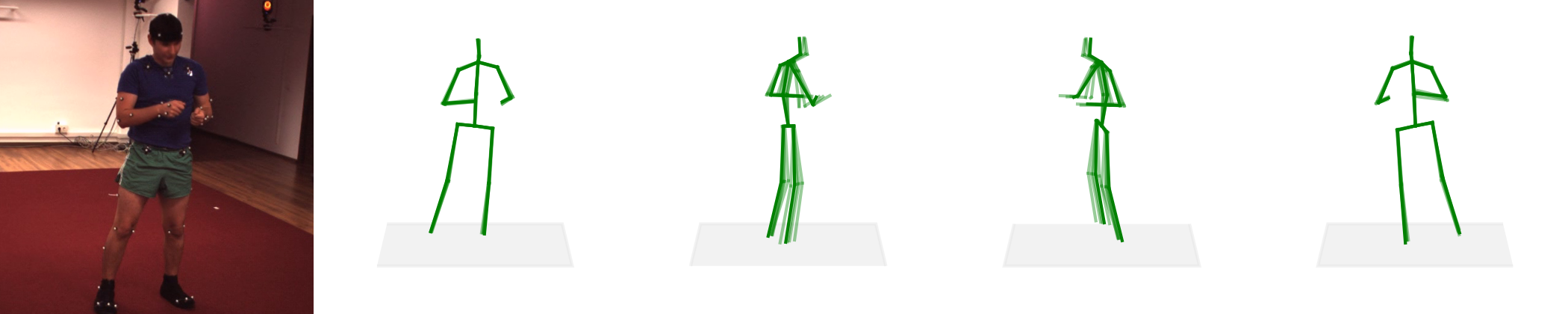}
    \includegraphics[width=\textwidth]{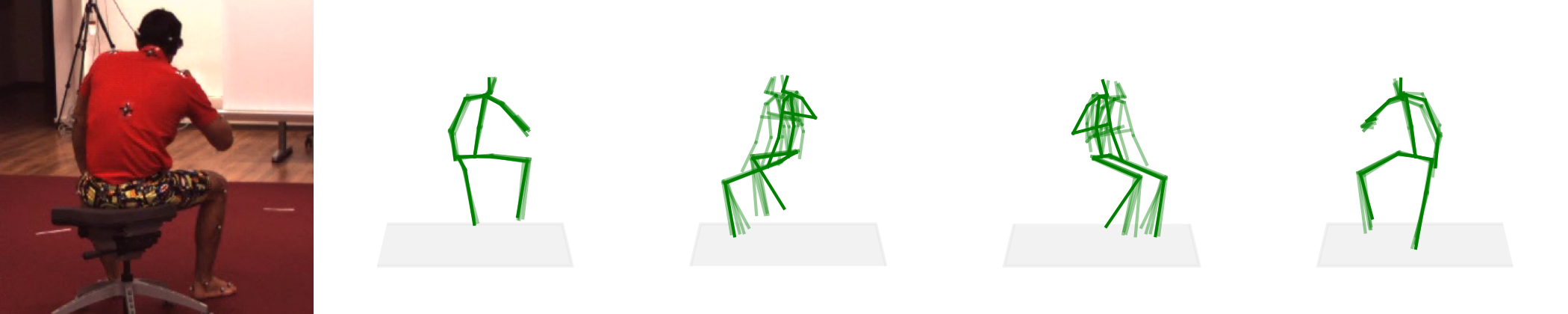}
    \includegraphics[width=\textwidth]{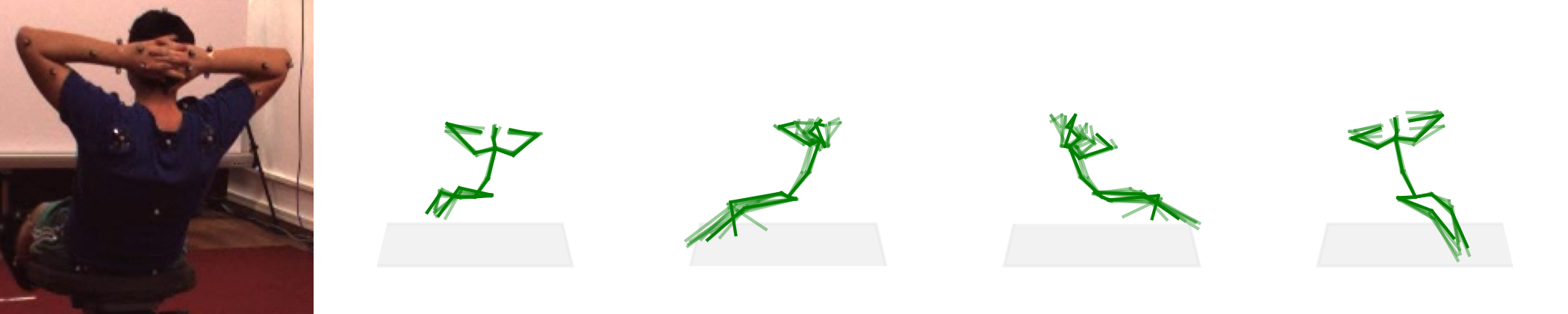}
    \includegraphics[width=\textwidth]{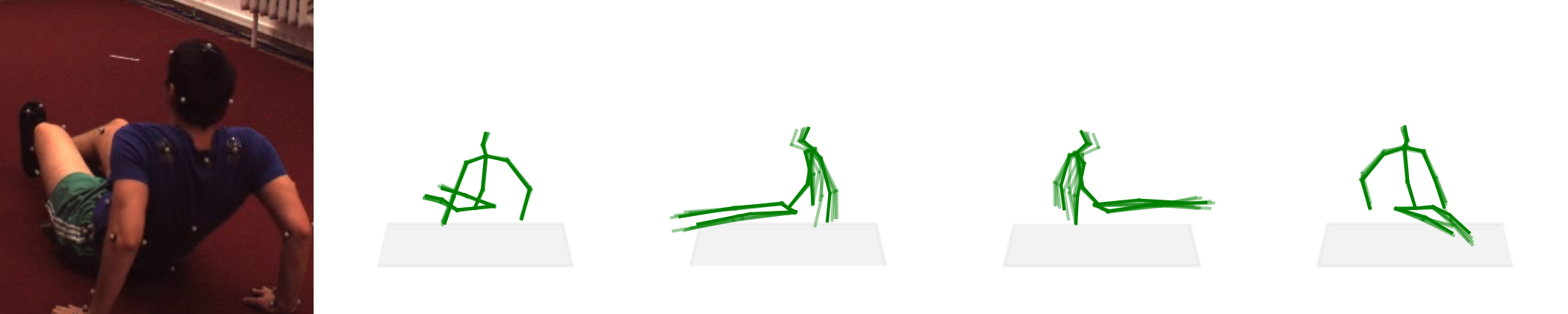}
    \includegraphics[width=\textwidth]{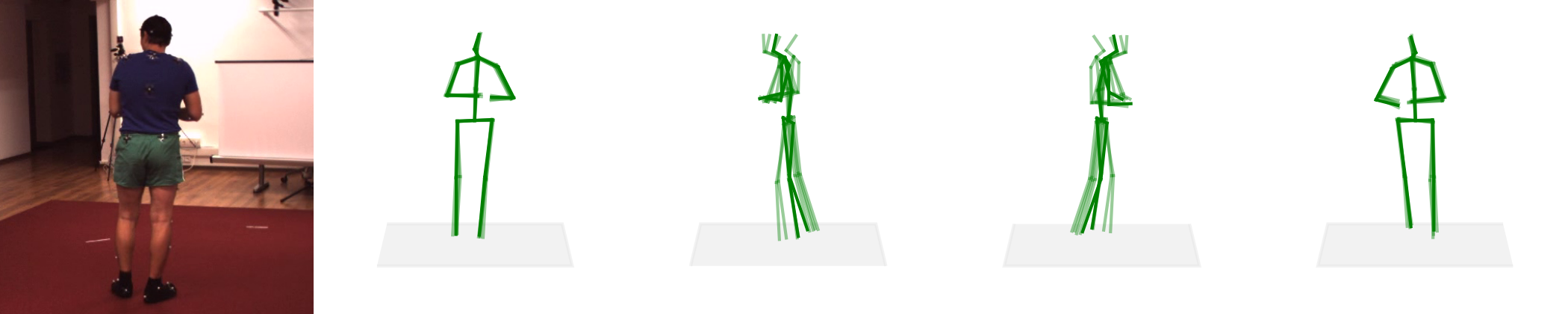}
    \caption{Examples for multiple hypothesis pose estimation. Poses are shown from the front, left, right and rear view points in each column. One sample is shown in bold, other samples are shown in a lighter shade for visual clarity. Note the higher variance in the side views (columns two and three) due to depth ambiguity.}
    \label{fig:multi_hypothesis_examples}
\end{figure*}

We present qualitative examples of multiple hypothesis pose estimation in \cref{fig:multi_hypothesis_examples}. Note the greater variance in the z (depth) dimension compared to the x and y dimensions. As our observation likelihood is defined in the image plane, the magnitude of gradient $\nabla_{\!x_t^{(j)}}\log p(c \mid x_t)$ is smaller in the z dimension (and is zero at the principal point), leading to greater variance.

\begin{figure*}
    \includegraphics[width=\textwidth]{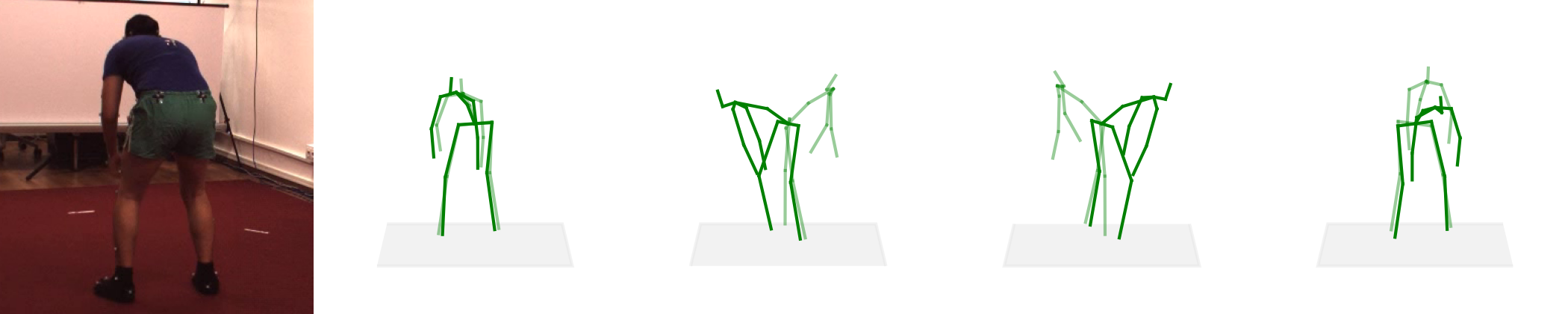}
    \includegraphics[width=\textwidth]{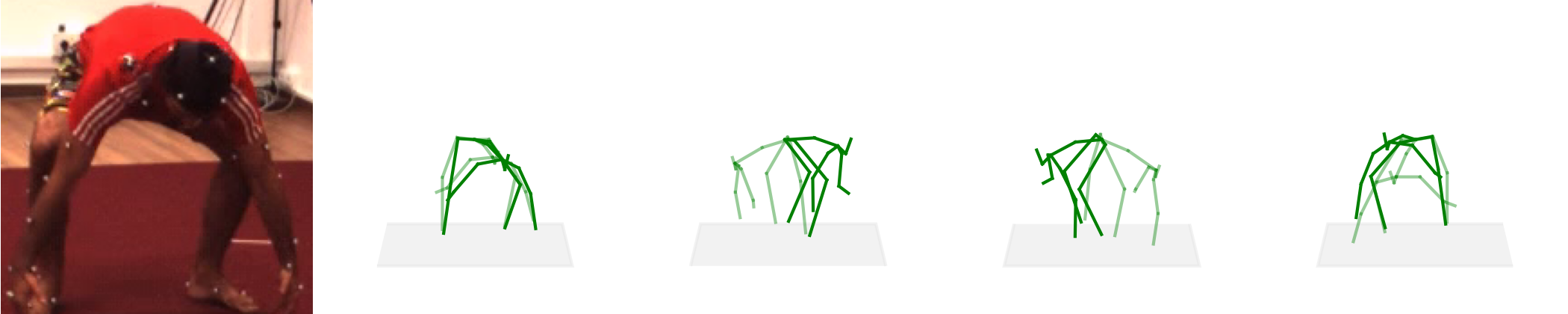}
    \includegraphics[width=\textwidth]{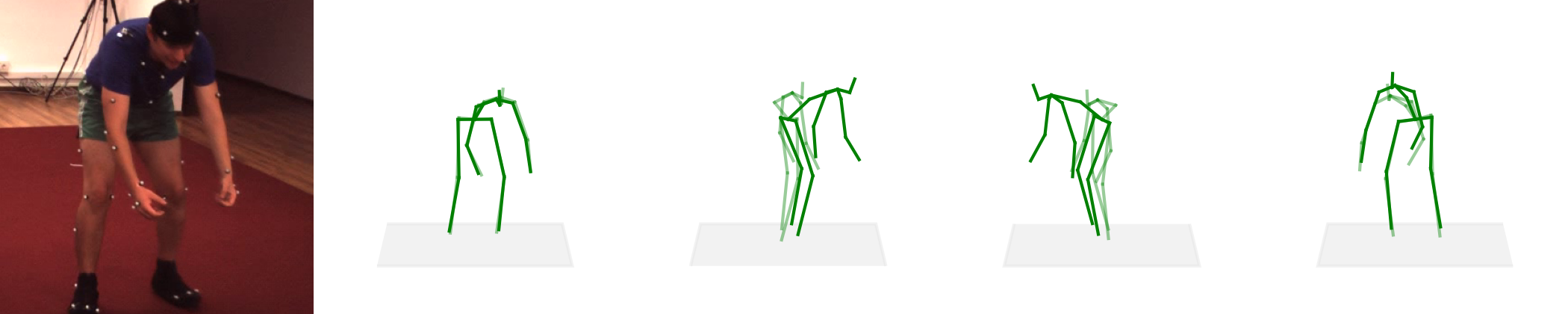}
    \caption{Failure cases for pose estimation. In particular poses with large rotation around the hips joints are problematic, with generated poses either rotating backwards in an implausible manner, or shortening the bone lengths to match the 2D detections. A plausible pose sample is shown in bold, with the failure case shown in a lighter shade.}
    \label{fig:failure_cases}
\end{figure*}

\textbf{Failure due to depth ambiguity:} There is inherent depth ambiguity in monocular methods, and while the increased diversity demonstrated above is a desirable capability, depth ambiguity can also causes a failure mode in our method. As the observation likelihood is based on re-projection to the image plane, there exists an infinite number of poses which match the 2D observations, however not all are plausible, and the observation likelihood can guide the reverse process into these kinematically implausible regions. This is particularly problematic for our method in cases with large depth variance, i.e., when the subject is bending towards or away from the camera. We present examples of these failure cases in \cref{fig:failure_cases}. Note that for each row, there are both correct and incorrect sampled poses, highlighting the benefit of estimating a probabilistic distribution instead of a single deterministic pose. The first column is from the camera view, and the second and third columns show the same samples from the side views where the failure mode is apparent. The first and second rows show incorrect poses where the upper body has rotated backwards instead of forwards around the hip joints. The third row is a case where a sample has implausible bone lengths, with the bones becoming elongated in the lower body, and compressed in the upper body. These poses maximize the observation likelihood as the projection of the 3D joints are consistent with the 2D detections, however are incorrect and kinematically implausible.

\section{Qualitative Pose Examples}
In this section we present additional qualitative pose examples for various tasks.

\begin{figure*}
    \includegraphics[width=\textwidth]{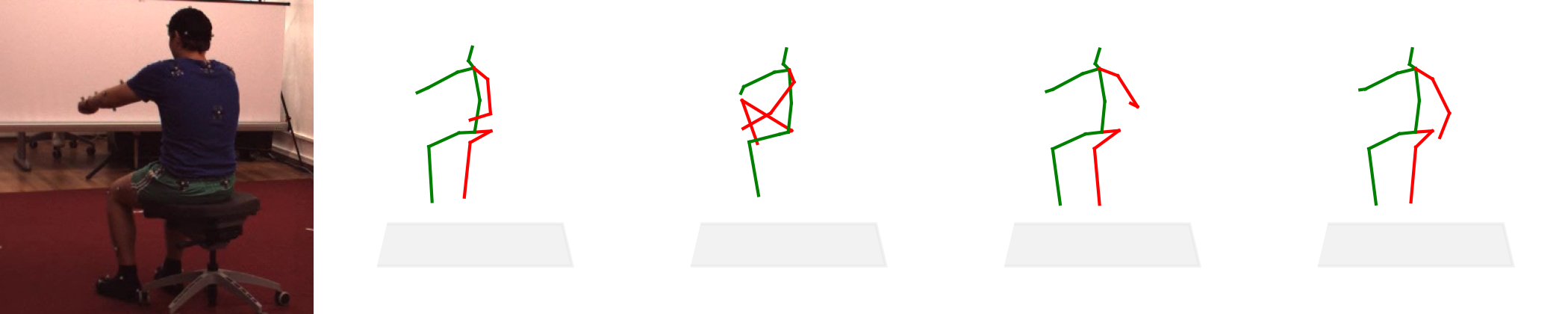}
    \includegraphics[width=\textwidth]{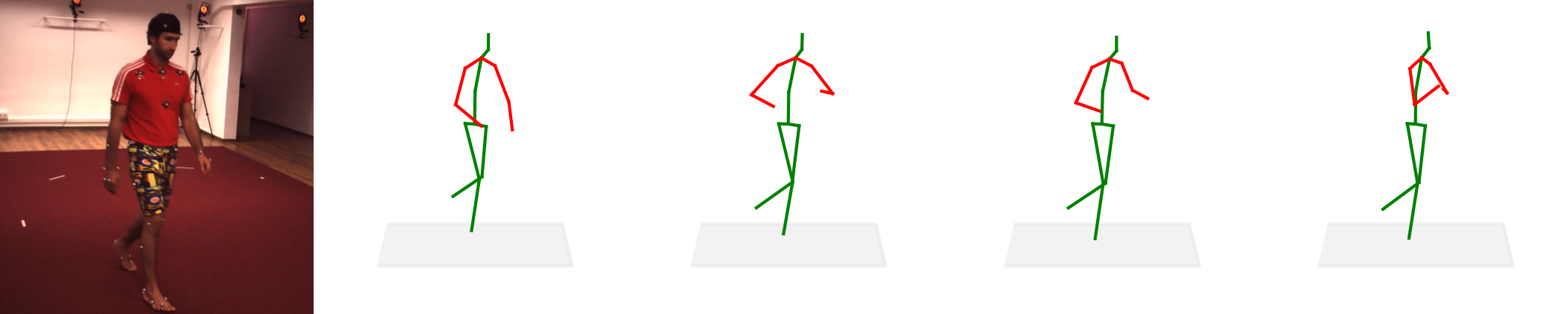}
    \includegraphics[width=\textwidth]{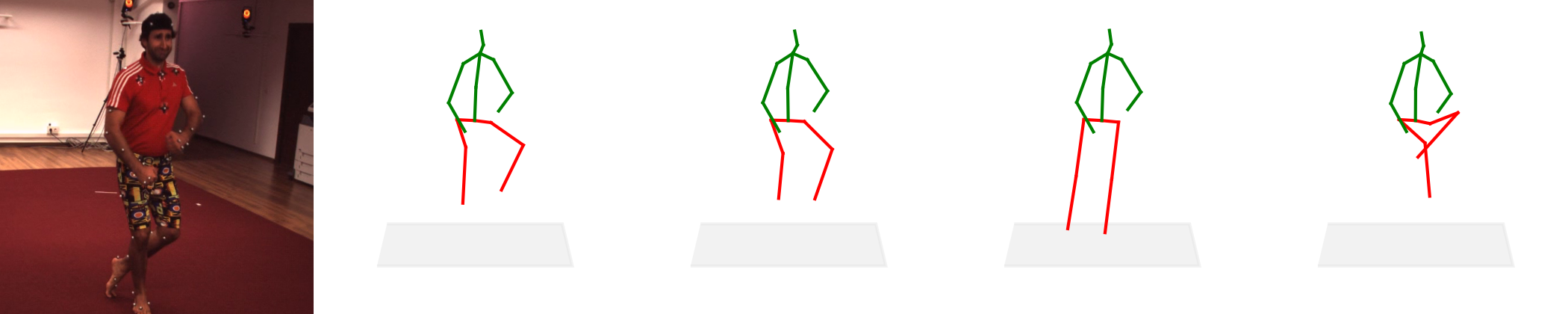}
    \caption{Pose completion example for different completion tasks. Red joints are inpainted by the pose prior.}
    \label{fig:pose_completion_examples}
\end{figure*}

\textbf{Pose Completion:} We present additional qualitative examples of pose completion in \cref{fig:pose_completion_examples}. Note that the sampled poses may not necessarily be consistent with the image due to the partial observation likelihood, however the samples are visually plausible as a complete human pose. The inpainted joints are coherent with the remaining joints, and bone lengths in limbs appear symmetric and of plausible length.

\begin{figure*}
    \includegraphics[width=\textwidth]{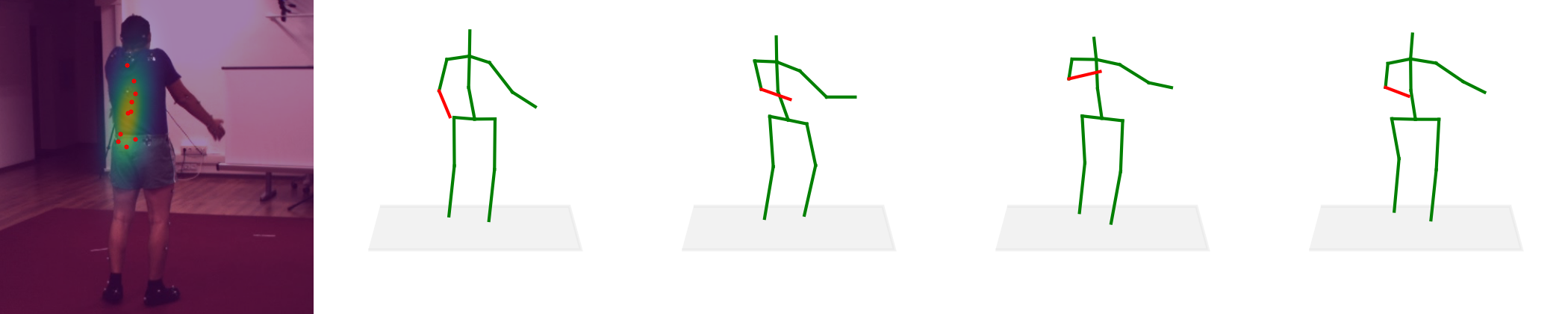}
    \includegraphics[width=\textwidth]{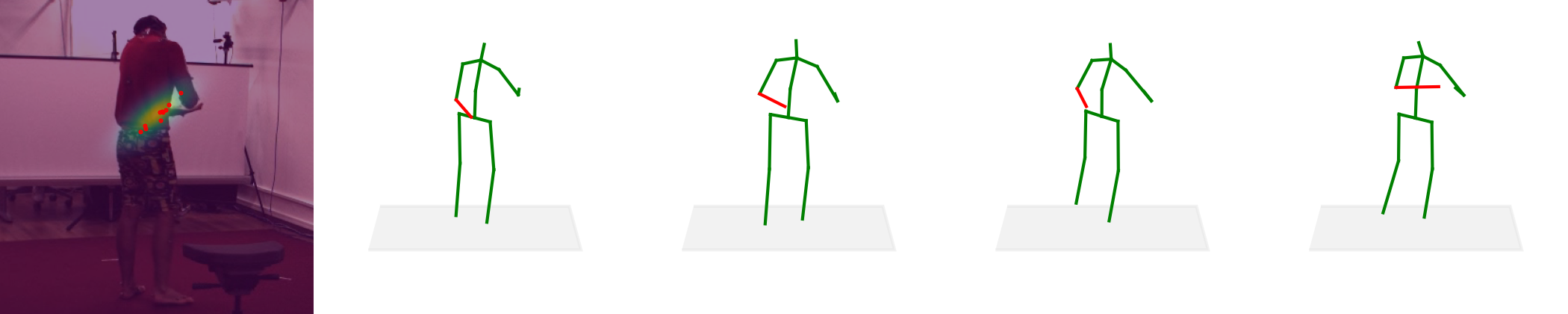}
    \includegraphics[width=\textwidth]{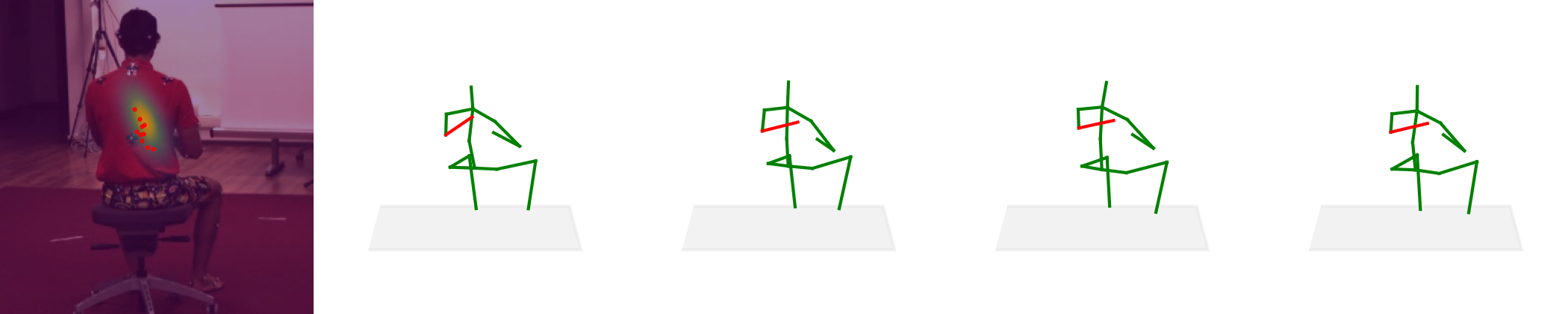}
    \caption{Examples showing pose estimation results for examples where 2D detections have high uncertainty. The observation likelihood and projection of the uncertain joints is shown on the image in column one, columns two-five show different pose samples. Projection of the uncertain joint is visually consistent with the covariance of the observation likelihood.}
    \label{fig:diversity_examples}
\end{figure*}

\textbf{Pose Diversity:} We show additional qualitative examples for 2D detections with high uncertainty in \cref{fig:diversity_examples}. Note that the projection of the uncertain joint is consistent with the covariance of the observation likelihood, indicating that the posterior pose distribution successfully reflects the uncertainty in the 2D detections.

\begin{figure*}
    \includegraphics[width=\textwidth]{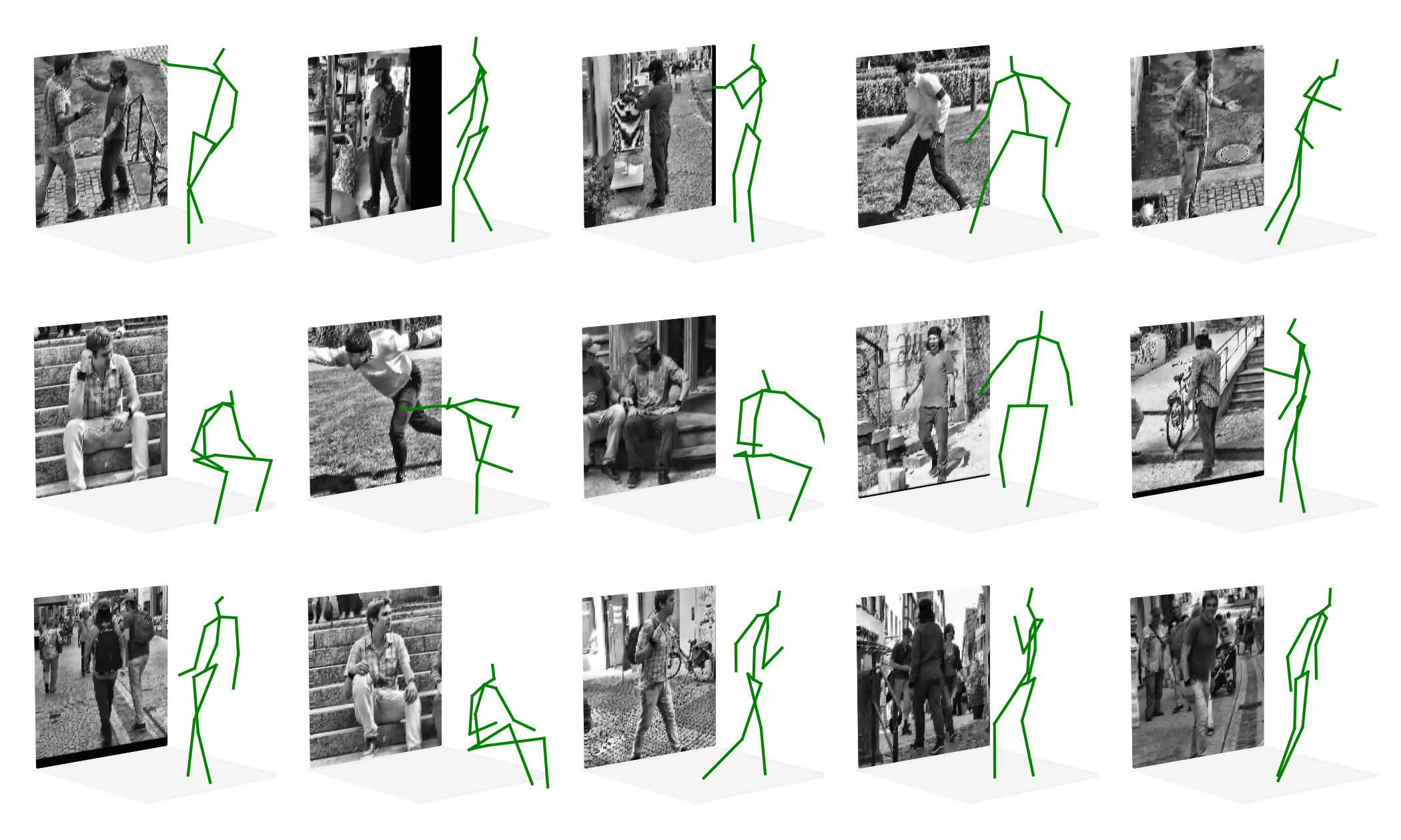}
    \caption{In-the-wild pose estimation examples from 3DPW dataset. Grayscale images illustrated for context. When multiple subjects are present, the pose result is for the subject centered in the image.}
    \label{fig:3dpw_examples}
\end{figure*}

\textbf{In-the-wild Examples:} The 3DPW dataset is particularly diverse, including in-the-wild scenes of multiple subjects. We present qualitative examples of pose estimation on 3DPW in \cref{fig:3dpw_examples}.
